# SYNONYM DETECTION USING SYNTACTIC DEPENDENCY AND NEURAL EMBEDDINGS

Dongqiang Yang, Pikun Wang, Xiaodong Sun and Ning Li
*School of Computer Science and Technology, Shandong Jianzhu University, Jinan 250101, China*

## ABSTRACT

Recent advances on the Vector Space Model have significantly improved some NLP applications such as neural machine translation and natural language generation. Although word co-occurrences in context have been widely used in counting-/predicting-based distributional models, the role of syntactic dependencies in deriving distributional semantics has not yet been thoroughly investigated. By comparing various Vector Space Models in detecting synonyms in TOEFL, we systematically study the salience of syntactic dependencies in accounting for distributional similarity. We separate syntactic dependencies into different groups according to their various grammatical roles and then use context-counting to construct their corresponding raw and SVD-compressed matrices. Moreover, using the same training hyperparameters and corpora, we study typical neural embeddings in the evaluation. We further study the effectiveness of injecting human-compiled semantic knowledge into neural embeddings on computing distributional similarity. Our results show that the syntactically conditioned contexts can interpret lexical semantics better than the unconditioned ones, whereas retrofitting neural embeddings with semantic knowledge can significantly improve synonym detection.

## KEYWORDS

Syntactic Dependencies, Distributional Semantics, Neural Network Embeddings

## 1. INTRODUCTION

Word meaning, represented in the distributional vector space model (VSM), usually employs co-occurrences in contexts, hypothesising similar words sharing similar contexts (Harris, 1985). To abstract distributional characteristics of words, Lowe (2001) proposes a VSM with quadruple operands (*B, A, S, M*), where (1) *B* consists of *basis elements* to form the dimensionality of a semantic space, which can be a group of words (Sahlgren, 2006), syntactic dependencies (Curran, 2003; Weeds, 2003), and the like; (2) *A* transforms raw co-occurrence frequencies between words and the *basis elements* using functions such as Pointwise Mutual Information (PMI) and log-likelihood ratio; (3) *S* stands for similarity methods predicting semantic similarity through distributional context similarity, which often includes the *cosine* similarity and *Euclidean distance*; (4) *M* performs dimensionality reduction on a semantic space through Singular Value Decomposition (SVD) (Schütze, 1992), Random Indexing (RI) (Kanerva et al., 2000), and the like. There are some comprehensive surveys on VSM in the literature. For example, Bullinaria and Levy (2006) mainly investigated the factors affecting distributional similarity in an unconditioned (a bag of words) setting such as the size of context window and similarity measures in the evaluations of multiple-choice synonym judgements, semantic and syntactic categorisation, and the like; Padó and Lapata (2007) compared the difference between syntactically conditioned (syntactic dependencies) or unconditioned dimensionality of VSM; Baroni et al. (2014) systematically compared the context-predicting neural embeddings with the traditional context-counting VSMs through many lexical semantics tasks.

However, the salience of syntactic dependencies in VSM has not been thoroughly investigated. Previous studies on the topic (Pantel and Lin, 2002; Curran, 2003; McCarthy et al., 2004) failed to distinguish nuances of grammatical relations and simply assembled different syntactic dependencies into one unified representation, similar to deducing distributional semantics with an unordered bag of words. Although Padó and Lapata (2007) attempted to investigate the role of each syntactic dependency in VSM through a predefined weighting scheme, they have not clearly shown to what extent one single type of syntactic dependency can contribute to distributional semantics. Observing that the syntactically conditioned representation in VSM can provide more helpful clues for distributional semantics (Hirschman et al., 1975; Hindle, 1990) than the unconditioned one, we focus on studying the salience of major types of grammatical relations in regulating distributional semantics. Note that we employ traditional corpus statistics rather than neural language models (Bengio et al., 2003) to investigate syntactically constrained VSM.

Neural network embeddings (NNEs), such as the unified (Mikolov et al., 2013b; Pennington et al., 2014) and contextualised ones (Devlin et al., 2018; Howard and Ruder, 2018; Peters et al., 2018; Radford et al., 2018), have achieved significant progress in the downstream applications of NLP. To thoroughly investigate the dependency constrained VSM in detecting synonyms, we also cover many typical neural embeddings such as Skip-gram with Negative Sampling (SGNS) (Mikolov et al., 2013a; Mikolov et al., 2013b), GloVe (Pennington et al., 2014), and fastText (Bojanowski et al., 2017). Moreover, we incorporate the knowledge enhanced NNEs that inject lexical semantic relationships into the distributional VSMs (Faruqui et al., 2015; Mrkšić et al., 2016). In addition to distributional semantics in synonym detection, we also employ knowledge-driven methods in the evaluation. Overall, in comparison with different counting/predicting word embeddings, along with taxonomic semantics, we systematically investigate syntactically constrained VSM in deriving distributional semantics.

## 2. SYNTACTICALLY CONSTRAINED VSM

To deduce distributional semantics with grammatical relations in VSM, we usually conduct the following procedures: (1) pre-processing sentences in the corpora with shallow/complete parsing; (2) extracting and/or categorising syntactic dependencies into distinctive subsets or vector spaces according to head-modifier (including adjective-noun and adverb or the nominal head in a prepositional phrase to verb) and grammatical roles (including subject-verb and object-verb); (3) applying the transformation of Singular Value Decomposition (SVD) (Schütze, 1992) on the dependency sets to create the latent semantic representations; and (4) determining distributional similarity using similarity measures such as the Jaccard coefficient and *cosine*, or probabilistic measures such as KL divergence and information radius.

## 2.1 Syntactic Dependency Contingency

Word sense disambiguation can leverage syntactic dependencies in context, where the semantic requirements are bi-directional in the form of head-modifier and head-complement (Cruse, 1986). As shown in Table 1, in the semantic traits of construction, the dominant role of the selector is prevalent in the determination requirement, which is also facilitated with the additional dependency requirements (Cruse, 1986). In one of Cruse's examples on head-modifier, e.g., *pregnant cousin*, the modifier *pregnant* dominates the female attributes of *cousin*, whereas *pregnant*, as the depender in the dependency restriction, also adds some features absent from *cousin*. To thoroughly study the syntactic dependencies in VSM, we mainly cover four types of grammatical relationships (i.e., **RV**, **AN**, **SV**, and **VO**), as listed in Table 1.

Table 1. The relation types in dependency construction

|  | *determination* | | *dependency* | |
|---|---|---|---|---|
|  | *selector* | *selectee* | *depender* | *dependee* |
| *constructions* | *modifier* | *head* | *modifier* | *head* |
|  | *head* | *complement* | *complement* | *head* |
| **RV** | verb modifiers:{adverbs|head nouns}→ verbs | | | |
| **AN** | Noun modifiers:{adjective| pre/post-modification}→nouns | | | |
| **SV** | {subjects}→{predicates} | | | |
| **VO** | {predicates}→{objects} | | | |

## 2.2 Syntactic Dependency Matrix

Following similar works on using syntactic parsers in distributional semantics, e.g., shallow parsers (Grefenstette, 1992; Curran, 2003) and a full parser MINIPAR (Lin, 1998), to collect distributional information, we propose to construct VSMs for the four types of syntactic dependencies through the Link Grammar parser (Sleator and Temperley, 1991). In a similar vein to Yang and Powers (2010) in automating thesaurus construction with grammatical relations, we employed the Link Grammar parser to analyse the 100 million-word British National Corpus (BNC). After filtering out non-content words and conducting morphology and lemmatisation pre-processing, we separately retrieved four types of grammatical relationships in Table 1 to construct four corresponding raw matrices, denoted as $X_{RAW}$: $X_{RV}$, $X_{AN}$, $X_{SV}$, and $X_{VO}$.

Consider $X_{SV}$ an *m* by *n* matrix representing subject-verb dependencies between *m* subjects and *n* verbs. We illustrate the **SV** relation using the rows ($X_{SV}$ or $\{X_{i,*}\}$) of $X_{SV}$ corresponding to nouns conditioned as subjects of verbs in sentences, and the columns ($X_{SV}$ or $\{X_{*,j}\}$) to verbs conditioned by nouns as subjects. The cell $X_{i,j}$ shows the frequency of the $i^{th}$ subject with the $j^{th}$ verb. The $i^{th}$ row $X_{i,*}$ of $X_{SV}$ is a profile of the $i^{th}$ subject in terms of its all verbs and the $j^{th}$ column $X_{*,j}$ of $X_{SV}$ profiles the $j^{th}$ verb *versus* its subjects.

Our matrices are very sparse, with zeros in over 95 percent of the entries. For each matrix, we transformed each cell frequency $freq(X_{i,j})$ into its information form using $log(freq(X_{i,j})+1)$ while retaining matrix sparsity. Apart from the logarithmic $freq(X_{i,j})$, Landauer and Dumais (1997) also divided it by the entropy of the column vector $X_{*,j}$ to adjust the association between words and documents, where the maximum of entropy occurred when every word in the row vector occurred evenly in one document, and the minimum is when one word is represented in the document. It can be formulated as:

$$freq(X_{i,j}) \Rightarrow \log(freq(X_{i,j})+1) \Big/ -\sum_{k=1}^{|X_{*,j}|} P(X_{k,j}) \log(P(X_{k,j}))$$

where $|X_{*,j}|$ is the size of $X_{*,j}$, and $P(X_{k,j})$ is the probability of the cell $X_{k,j}$ to the sum of $X_{*,j}$. The entropy in this formula functions similarly to the Inverse Document Frequency (IDF) to the Term Frequency (TF) in IR. As an alternative, Rapp (2003) proposed to multiply by the entropy in calculating distributional similarity for clustering word senses, which was based on a word by word association matrix. Landauer and Dumais (1997) and Rapp (2003) employed the context of a bag of words, say, word co-occurrences in a fixed-size window, so that data sparseness was not as severe as in our syntactic dependency matrices. We did not apply the dampening factor of IDF-like entropy on co-occurrences acquired under the condition of syntactic dependencies as our matrices are much sparser than a typical bag-of-word VSM, and it would be superfluous to regularise the matrices repeatedly.

## 2.3 Finding Principal Components

To further reduce the dimensionality of these matrices, we applied SVD/LSA (Deerwester et al., 1990; Landauer and Dumais, 1997) on them to transform syntactically constrained VSMs into their latent semantic space models. In the investigation of using LSA to find synonyms, Landauer and Dumais (1997) claimed that the optimal performance was subject to variation on the number of single values or principal components. The components in the compressed space reflect mainly semantic features, attributes, or concepts that are reminiscent of the human semantic memory model (Quillian, 1968). In Roget's Thesaurus ver. 1911, there are nearly 1,000 semantic categories, which organise over 40,000 words. We fixed 1,000 as the default size of each word vector in the semantic space and reduced all matrices to 1,000 singular values or eigenvectors concerning the expensive computation of SVD on these sparse matrices. To further select the appropriate number of singular values out of 1,000, we defined the selection probability $P_i$ of the single value $S_i$ to the relative variance it can stand for, $P_i = S_i^2/\Sigma(diag(S^2))$, where $\Sigma(diag(S^2))$ is the sum of squares of 1,000 singular values. Among the singular values, the first 20 components account for around 50% of the variance, and the first 250 components for over 75%. We established 250 as a fixed size of the compressed semantic space. In the following sections we will denote the syntactically conditioned co-occurrences or raw matrices as $X_{RAW}$, in contrast to the SVD compressed ones $X_{SVD}$.

To measure the effectiveness of syntactically-conditioned $X_{RAW}$ and $X_{SVD}$ on mining latent semantic components, we employ the *cosine* similarity of word vectors as used in LSA and commonly adopted in assessing distributional similarity.

## 3. MULTIPLE-CHOICE SYNONYM JUDGEMENTS

Landauer and Dumais (1997) evaluated SVD/LSA in lexical knowledge acquisition through the synonym test part of TOEFL, in which each examinee was presented with 80 questions designed for assessing his/her ability in standard written English. Each question comprises a target word followed by its four alternative words, one of which is the semantically closest answer or synonym to the question word. On average, people from non-English speaking countries achieved 51.6 correct answers or a 64.5% correct rate, which was taken as a baseline for this task by Landauer and Dumais (1997). Since TOEFL benchmarks the evaluation of distributional similarity (Landauer and Dumais, 1997; Rapp, 2003, 2004; Bullinaria and Levy, 2006; Padó and Lapata, 2007), we employ it to evaluate the salience of the four syntactic dependency sets and their corresponding compressed counterparts after SVD regarding semantic knowledge acquisition.

### 3.1 A walk-through example

To demonstrate the quality of our syntactically constrained VSMs, we listed the top 10 similar words for verbs and nouns after calculating and ranking their distributional similarity (*cosine*) respectively in the SVD-compressed $X_{RV}$ and $X_{AN}$, as shown in Table 2. We selected a target verb or noun (in bold) with its frequency in BNC ranging from over 10,000 times (high frequency), between 10,000 and 4,000 times (medium frequency), and below 4,000 times (low frequency).

Table 2. Top 10 similar words after computing distributional similarity in $X_{SVD}$

|  | $X_{RV}$ | $X_{AN}$ |
|---|---|---|
| High frequency | **drink**: *sip pour slop bottle spill swill slurp smoke decant eat* | **branch**: *tree twig department bureau college shrub leaf faculty outlet institute* |
| Medium frequency | **decline**: *decrease dwindle deteriorate diminish shrink expand wane refuse multiply wither* | **recession**: *slump downturn drought crisis depression inflation boom upheaval shortage unemployment* |
| Low frequency | **deter**: *discourage penalize punish dupe tempt restrain justify coerce constrain nerve* | **jewel**: *necklace sari jewellery silver brooch scarlet livery scarf diamond braces* |

### 3.2 Synonym Detection

In line with Yang and Powers (2006), we first manually divided the 80 questions into four sub-question sets according to common PoS tags between a target or question word and its options or answers. They were evenly distributed on Part-of-Speech (PoS) tags, namely 23 adjectives (29%), 20 verbs (25%), 19 nouns (24%), 18 adverbs (23%). Secondly, we selected the option word with the highest similarity of the target word for each subset of the same PoS, after separately calculating the *cosine* similarity of the target word and its one of four options within XRAW and XSVD correct answer or synonym to the target. Thirdly, to finalise the total number of answers found in $X_{RAW}$ and $X_{SVD}$, we avoided simply averaging the sum of the number of answers $X_{AN}$, $X_{RV}$, $X_{SV}$, and $X_{VO}$ in each sub-question set. Rather, we calculated $Ans_i$, the answer score to the $i^{th}$ question $Que_i$ (the target word), to be equal to 1, indicating a correct answer to $Que_i$, if and only if (1) $TF_{i,m}$, the term frequency (*TF*) of $Que_i$, holds maximum in the $m^{th}$ matrix of {$X_{An}$, $X_{aN}$, $X_{Rv}$, $X_{rV}$, $X_{Sv}$, $X_{sV}$, $X_{Vo}$, and $X_{vO}$} and (2) $Ans_{i,m}$, the answer score to $Que_i$ in the $m^{th}$ matrix is also equal to 1; otherwise $Ans_i = 0$. This can be formulated as follows: $Ans_i = 1$, if $Max(TF_{i,m}) > 0$ and $Ans_{i,m} = 1$; otherwise $Ans_i = 0$. This function implies that the final correct answer to each question is credited if there is a correct answer in the matrix where the term frequency of the question word is maximum across all matrices. The total number of correct answers in $X_{RAW}$ and $X_{SVD}$ that distributional similarity (*cosine*) can work out is the sum of $Ans_{i,m}$ across the four sub-question sets.

Consider 1 of 80 questions in TOEFL: finding the synonym of *hasten* from a group of words including *accelerate, permit, determine,* and *accompany,* for example. The distributional similarity (*cosine*) was first calculated in each subset, as shown in Table 3, where the figure in parentheses is the term frequency of *hasten* in each raw co-occurrence matrix.

Table 3. Finding a synonym of the verb *hasten* in the English synonym test of TOEFL

| *hasten* | $X_{rV}$ (226) | | $X_{Vo}$ (235) | | $X_{sV}$ (148) | |
|---|---|---|---|---|---|---|
| | $X_{RAW}$ | $X_{SVD}$ | $X_{RAW}$ | $X_{SVD}$ | $X_{RAW}$ | $X_{SVD}$ |
| *accelerate* | 0.070 | **0.247** | **0.620** | **0.606** | 0.143 | 0.353 |
| *permit* | 0.082 | 0.071 | 0.123 | 0.109 | 0.102 | 0.086 |
| *determine* | **0.085** | -0.047 | 0.073 | -0.041 | 0.107 | -0.014 |
| *accompany* | 0.061 | 0.049 | 0.195 | 0.365 | **0.176** | **0.383** |

Note that there is no occurrence of *hasten* as a noun in $X_{AN}$. Furthermore, the word *hasten* occurred 235 times with its objects in $X_{Vo}$, more frequently than with its modifiers in $X_{rV}$ (226) and subjects in $X_{sV}$ (148), whereas *accelerate* had the highest distributional similarity with *hasten* in the raw co-occurrence matrix $X_{RAW}$ and the compressed matrices $X_{SVD}$. So, *accelerate*, the synonym of *hasten*, is the correct response on the assumption that the higher distributional similarity among words implies the closer semantic similarity.

## 3.3 Answer Distribution Under Different PoS Tags

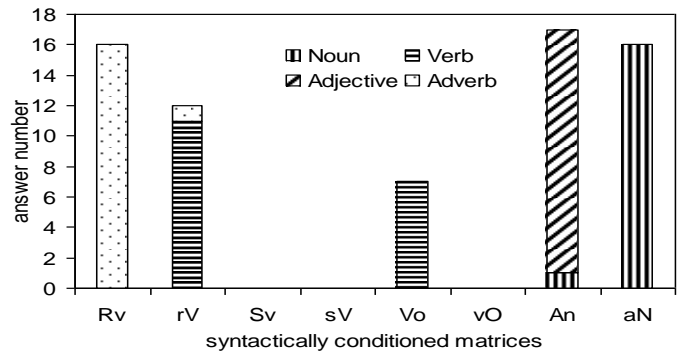

Figure 1. The total number of answers found in each dependency matrix

We correctly addressed 58 and 68 out of the 80 questions, respectively, using $X_{RAW}$ and $X_{SVD}$ in the TOEFL test. In the 68 correct answers from $X_{SVD}$, nearly all adverb questions (94%) were elicited correctly from $X_{Rv}$, where $X_{RV}$ is the syntactic set containing the relations between verbs and their modifiers, as shown in Figure 1. All correct adjective answers came from $X_{An}$ that mainly specifies modifiers such as adjectives with the modified nouns. Almost all correct noun answers were achieved in $X_{aN}$, which indicates their corresponding modifiers specifying the dominant distributional features of nouns. The total number of correct verb answers was 11 in $X_{rV}$ and 7 in $X_{Vo}$, respectively. It implies that both the modifiers and objects of the verbs could affect the semantic prediction of the verbs. Overall, the modifiers of nouns are likely to play an important role in accounting for their semantic features through distributional similarity; and most semantic features of verbs depend on their modifiers consisting mainly of adverbs, head nouns in the prepositional phrases, along with their objects.

## 3.4 Comparison to Counting-Based VSMs

The synonym test of TOEFL is widely adopted in evaluating statistical semantics acquired in the syntactically conditioned and unconditioned VSMs (Bullinaria and Levy, 2006; Padó and Lapata, 2007). We further compared distributional similarity (*cosine*) on the $X_{RAW}$ and $X_{SVD}$ ($cos_{RAW}$ and $cos_{SVD}$) with other state-of-the-art statistics-counting methods in this test, as shown in Figure 2.

For these comparisons, the baseline of people taking English as the Second Language (BL-ESL) denotes the average level of non-native English speakers in the test, containing 51.6 correct answers. The methods using unconditioned word co-occurrences mainly include:

- LSA: Landauer and Dumais (1997) first created a word-by-document matrix (60,768 by 30,473) from an encyclopedia of 4.6 million words, which was then normalised with logarithms where each cell frequency was divided by the entropy of a word across all its documents. To find synonyms, *cosine* on the SVD compressed matrix (reduced to 300 dimensions) achieved 64.4% accuracy on TOEFL.
- PMI-IR: Turney (2001) used the Pointwise Mutual Information-Information Retrieval (PMI-IR) algorithm that mainly retrieved word occurrences within a 10-word window through the *NEAR* query from the Alta Vista search engine and then calculated word association strength with mutual information to predict answers. 73.8% of TOEFL questions were correctly addressed.
- LC-IR: Higgins (2004) proposed a similar algorithm to PMI-IR, Local Context Information Retrieval (LC-IR), using the Alta Vista search engine. Instead of the *NEAR* query that retrieves word co-occurrences in a ±10 window, he collected words adjacent to each other within one-word distance to compute word distributional similarity. LC-IR reached 81.3% accuracy on TOEFL.
- Paradig: In investigating the word space model, Sahlgren (2006) divided plain contexts into syntagmatic and paradigmatic. The syntagmatic context provides word association for computing distributional similarity, which holds a similar assumption to PMI-IR and LC-IR. In contrast, the paradigmatic representation focuses on word interchangeability in the same environment of surrounding words, which is analogous to LSA. With two different versions of corpora, BNC and the Touchstone Applied Science Associates (TASA), comprising 10 million words and a collection of high-school level English texts over many topics such as science and health, his results in TOEFL showed a more significant number of correct answers was acquired using the paradigmatic context (75% accuracy in TASA vs 72.5% in BNC) than the syntagmatic one (67.5% in BNC vs 52.5% in TASA).
- Rapp: Rapp (2003) reported an excellent result of 92.5% accuracy in TOEFL. Apart from lemmatisation and functional words filtration of BNC and SVD reduction on the dimensionality of semantic space (300 dimensions), he also normalised word co-occurrence using entropy-based transformation and removed lemmas with frequencies less than 20 times, which further lessened data sparseness and lowered the amount of noise in VSM. Without SVD, his approach using *cosine* arrived at 69% accuracy.

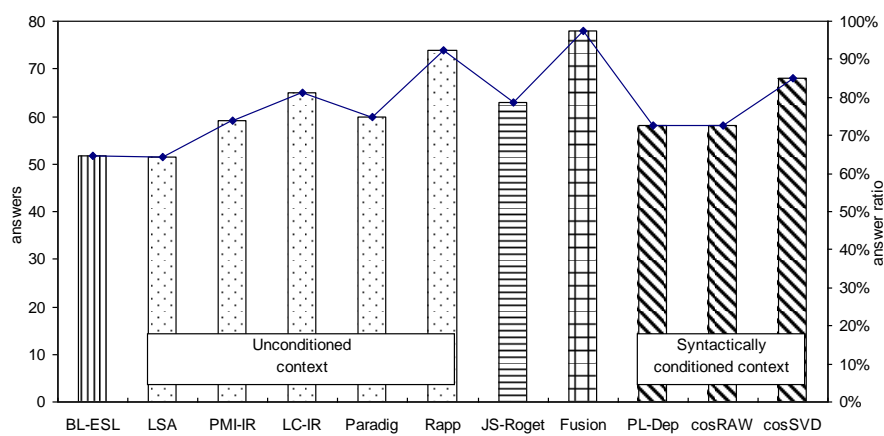

Figure 2. A performance comparison of different methods in the English synonym test of TOEFL

For the VSM conditioned on syntactic dependencies, Padó and Lapata (2007) (PL-Dep) implemented a platform for comparing different syntactic dependencies, normalisation methods, and similarity methods. Their optimal dependency-based model features weighting dependencies, concatenating no more than three dependencies, and mapping and reducing word dimensions into 2,000 basic ones. They correctly address about 73.0% of TOEFL questions using the information-theoretic similarity measure (Lin, 1998).

In addition to these methods using the syntactically conditioned and unconditioned contexts, Turney et al. (2004) proposed a fusion scheme (Fusion) based on the product rule, merging LSA, PMI-IR, Roget's Thesaurus based method (Jarmasz and Szpakowicz, 2003) (JS-Roget), and Connector using summary snapshots from Google querying (a similar technique to PMI-IR). Note that instead of computing distributional similarity, JS-Roget measured taxonomic similarity using the simple edge-counting of Roget's Thesaurus. Fusion achieved 97.5% accuracy on the 80 TOEFL questions, but it is not clear that is a generalisable approach. Note that in Figure 2, the methods using unconditioned context include LSA, PMI-IR, LC-IR, Paradig, and Rapp, whereas PL-Dep, $cos_{RAW}$, and $cos_{SVD}$ employ syntactically conditioned context.

The synonym test of TOEFL is designed to examine the lexical knowledge of non-native English speakers, and it was no surprise that JS-Roget outperformed LSA, PMI-IR, and LC-IR due to the effective organisation of Roget's Thesaurus. The reason that it failed to reach 100% of the correct answers is probably attributed to the lexical knowledge coverage of the machine-readable Roget in the experiment. The improvement of PMI-IR and LC-IR over LSA could be attributed to using terabyte-sized corpora that significantly reduced data sparseness in extracting synonyms. In contrast to LSA (64.4%), except for the same similarity measure (*cosine*), our competitive performance, $cos_{SVD}$ (85.0%) and $cos_{RAW}$ (72.5%), could be partly due to the word representation with syntactic dependencies rather than plain word co-occurrences in LSA, and partly due to the division of the syntactic dependencies into the four different subsets dedicated on one of the four dependencies.

Although the correctness of $cos_{SVD}$ (85.0%) and $cos_{RAW}$ (72.5%) in addressing the multiple-choice synonym questions is lower than Rapp, it is worth pointing out that without SVD reduction, Rapp only arrived at 69% accuracy. Therefore, the SVD compression on word co-occurrence space contributed much more in Rapp than $cos_{RAW}$ to $cos_{SVD}$. Note that Rapp filtered out lemmas with a frequency less than 20, and we did not set up a threshold to filter out the low-frequency cells. We assumed that all triples for the syntactic dependency yielded from the parser could contribute to the prediction of semantic similarity through distributional similarity, even if the parse was formally incorrect or represented an unlikely reading in the context.

On the other hand, Bullinaria and Levy (2006) conclude that corpus size and quality are essential factors in deciding the performance of VSMs. More reliable statistics of word co-occurrences can be derived from more prominent and better corpora with less unusual and noisy words. In their systematic experiments of distributional similarity on VSMs, Padó and Lapata (2007) observed that the VSM based on syntactically conditioned co-occurrences had significantly outperformed the VSM based on plain co-occurrences. Since our raw co-occurrence matrices, as a collection of word dependencies, are much sparser than simply counting neighbouring words in a $\pm n$ window without filtering out low-frequency word co-occurrences, there is still room to improve our syntactically conditioned VSM.

There is no doubt that the Fusion model that combines distinctive techniques has almost perfectly tackled the synonym test of TOEFL. Due to the encouraging results of $cos_{SVD}$, our method might be another vital resource to be merged with those in Fusion, which are mainly based on unconditioned contexts.

## 4. NEURAL EMBEDDINGS

Apart from counting-based VSM, we further contrast our dependency-constrained VSM with neural network embeddings, which have become a mainstream data representation for cutting-edge NLP applications such as machine translation and question answering. Instead of counting co-occurrence statistics in context to vectorise a word explicitly, neural language model (Bengio et al., 2003; Collobert and Weston, 2008) can implicitly generate word embeddings through either Skip-gram, maximising the probability of predicting its surrounding words for a central word, or CBOW, and vice versa (Mikolov et al., 2013a; Mikolov et al., 2013b). NNEs achieved state-of-the-art results in the benchmark evaluations of GLUE (Wang et al., 2018) for natural language understanding.

## 4.1 Distributionally Similarity

In contrast with the dependency-constrained VSM in Section 2, we employ three typical unified NNEs (i.e., representing different senses of a word with a uniform representation): the word2vec Skip-gram with Negative Sampling (SGNS) (Mikolov et al., 2013a; Mikolov et al., 2013b), GloVe (Pennington et al., 2014), and fastText (Bojanowski et al., 2017). To lessen the impacts of different hyperparameters, corpora, and text pre-processing procedures on training neural language models, SGNS (SN), GloVe (GV), and fastText (fT) were all trained with the English Wikipedia Dump of February 2017 (about 2 billion words) in a window of 5 words. Each token was also lemmatised before training (Fares et al., 2017). They are all featured with the same dimensionality of 300.

Given that distributional semantics induced from VSMs only specify word usage patterns in context, lacking human-compiled semantic knowledge, we, therefore, include ConceptNet Numberbatch (Speer and Chin, 2016), a hybrid NNE that retrofits VSMs with lexical knowledge bases (Faruqui et al., 2015). ConceptNet Numberbatch (CN in short) constructs a hybrid NNE through injecting ConceptNet 5.5 (Speer et al., 2017) into SN and GV, which were respectively pre-trained with the Google News dataset (about 100 billion words) and the Common Crawl (about 840 billion words).

Table 4. Top 10 distributionally similar words after computing *cosine* similarity in neural embeddings. A word is colorized if it appears at least in 3 of the NNEs

| | | |
|---|---|---|
| **drink** | SN | *beverage* *drinking* *beer* *drunk* eat alcoholic bottle consume alcohol food caffeinate juice |
| | GV | *beverage* non-alcoholic *beer* non-carbonated lemonade carbonated *drinking* *drunk* caffeinate juice |
| | FT | *beverage* non-alcoholic *drinking* lemonade *drunk* caffeinate *beer* carbonated alcoholic decaffeinate |
| | CN | whiskey_neat drinkless thirsty_person consume_beverage whiskey_on_rocks bedrink quench_thirst epotation in_drink bibacity |
| **branch** | SN | *Branch* stem line connection connect trunk originate closely root extend |
| | GV | sub-branch *Branch* Branches dichotomously offshoot bifurcate Nyungan unbranched Polygalaceae genicular |
| | FT | sub-branch multi-branched *Branch* Branches unbranched branchlet offshoot bifurcate anabranch branchial |
| | CN | branchlike part_tree branch_temple tree_part on_tree ramiform ramulose arborize underbranch ramify |
| **decline** | SN | decade *increase* *decrease* dramatically popularity however rise recent steadily due |
| | GV | precipitously dwindle stagnate downturn plummet wane *decrease* *increase* diminish upswing |
| | FT | dwindle stagnate precipitously wane *decrease* steadily downturn diminish dramatically *increase* |
| | CN | declines on_wane downward_trend degringolade declensionist detrect decliner declining falloff postphilosophical |
| **recession** | SN | *downturn* *late-2000s* *crisis* *Depression* *slowdown* unemployment slump economic economy inflation |
| | GV | *downturn* *late-2000s* *slowdown* *crisis* stagflation deflation deindustrialization slump *Depression* stagnation |
| | FT | *downturn* *late-2000s* stagflation deflation *slowdown* *crisis* debt-to-gdp *Depression* recessional late-2000 |
| | CN | economic_downturn recessionary great_recession downturn antirecession economic_condition megarecession wage_setter relicted downspin |
| **deter** | SN | *discourage* *deterrent* *dissuade* *prevent* *harm* aggression threat repel aggressor forestall |
| | GV | *prevent* *discourage* *deterrent* *dissuade* forestall avoid thwart *harm* impede protect |
| | FT | *deterrent* *discourage* *prevent* deterrence *dissuade* aggressor *harm* counterproductive warn threat |
| | CN | *dissuade* deterring *discourage* dehort *prevent* becourage dehortatory dissuades deterred dispurpose |
| **jewel** | SN | *precious* *necklace* jewelry treasure *gem* jewellery pearl diamond *ruby* priceless |
| | GV | *necklace* jeweled *ruby* pearl treasure earring priceless *gem* diadem *precious* |
| | FT | jeweled *necklace* jewelled jewelery *precious* earring jewelry jewellery *ruby* brooch |
| | CN | unjewelled unbejewelled foil_stone *precious*_stone *gem* gemmated gems jewels bejewel gemmed |

To demonstrate the quality of these NNEs in inferring distributional semantics, we retrieved the top 10 words after calculating *cosine* similarity between a target word and all the other words in the vocabulary, as listed in Table 4. The target words have a varied range of frequencies from high to low, the same as in Table 2. Results in Table 4 indicated that CN has a different latent semantic space from the native NNEs and shares a small number of common words with SN, GV, or fT, especially when dealing with high or medium frequency ranges. Although SN, GV, and fT exercise different training policies, they show similar patterns in deducing semantic similarity. For example, SN, GV, and fT rank *beverage* and *downturn* as the most similar words to *drink* and *recession*.

The results from the dependency-constrained VSM, as shown in Table 2, are distinctive from them in Table 4. Our proposal can retrieve semantically similar words with the same PoS tags in computing distributional similarity, whereas NNEs can yield semantically associated words that inevitably aggregate different PoS tags. For example, for the target verb *drink* in Table 2, its top 10 similar words all belong to the same PoS tag (i.e., verb). However, in Table 4, the top 10 similar words to *drink* contain *beverage* and *beer*, sharing the noun PoS tag, which are only semantically associated to *drink* as a verb in NNEs.

NNEs often yield distributionally similar words that are not necessarily semantically similar rather than associated, and sometimes they are even antonymous (Hill et al., 2015; Yang and Yin, 2021). For example, both *decrease* and *increase* appeared in the top 10 similar words of *decline* in SN, GV, and fT in Table 4.

## 4.2 Thesaurus Lookup

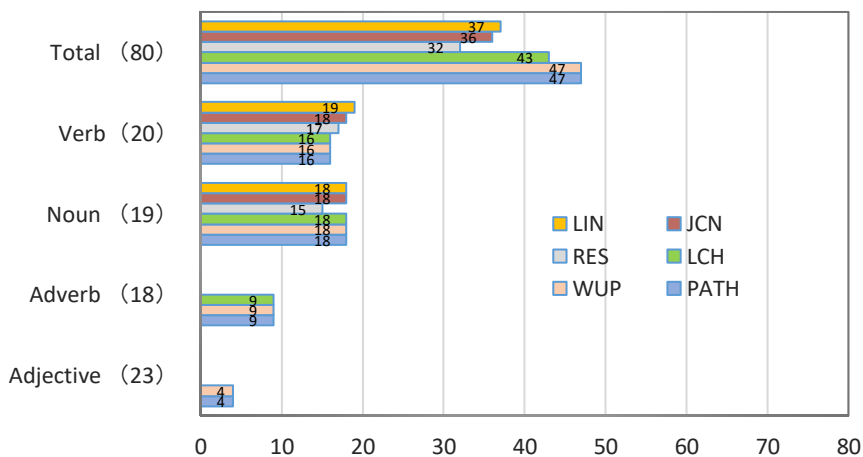

Figure 3. Comparison of different WordNet-based methods in detecting synonyms of TOEFL. The number of synonym questions under each subgroup is enclosed in the bracket beside each PoS tag

Before using distributional semantics to detect synonyms, we first set up a baseline that mainly uses human-compiled semantic knowledge to search for synonymous relations. We employ an NLTK (Bird and Loper, 2004) package for computing lexical semantic similarity (Pedersen et al., 2004) on WordNet (Miller et al., 1990; Fellbaum, 1998) to detect synonyms in TOEFL.

In the similarity package, PATH (Rada et al., 1989), WUP (Wu and Palmer, 1994), and LCH (Leacock and Chodorow, 1994, 1998) are based on calculating the shortest path length to derive taxonomic similarity, whereas RES (Resnik, 1995), JCN (Jiang and Conrath, 1997), and LIN (Lin, 1997) also factor in the statistics of word senses to modify the universal distance of path length in a taxonomy. These methods can calculate taxonomic similarity with their respectively designed semantic networks in WordNet for nouns and verbs, as shown in Figure 3. However, RES, JCN, and LIN failed to find synonyms for adjectives and adverbs that are not organised in a taxonomy. The path-length methods, including PATH, WUP, and LCH, can work on adjectives and adverbs by retrieving synonymous relationships. The best results in TOEFL were achieved by PATH and WUP, and both of them can identify 47 correct answers in total.

## 4.3 Synonym Detection

Like the answer finder in Section 3.2, we also calculate the *cosine* similarity between a question word and its four candidates. Since NNEs are stored only in a unified matrix rather than in different ones like our dependency-constrained VSM, we hypothesise that the correct answer to the question word should have a maximum similarity score. We illustrated the results for the synonym test in TOEFL in Figure 4, where CN can find 79 correct answers in total and significantly outperformed SN, GV, and fT. CN also achieved better results under each PoS category than the other 3 NNEs. Since CN is a hybrid NNE that retrofits distributional semantics with semantic knowledge, it is no wonder that it can surpass other NNEs and our dependency-constrained VSM.

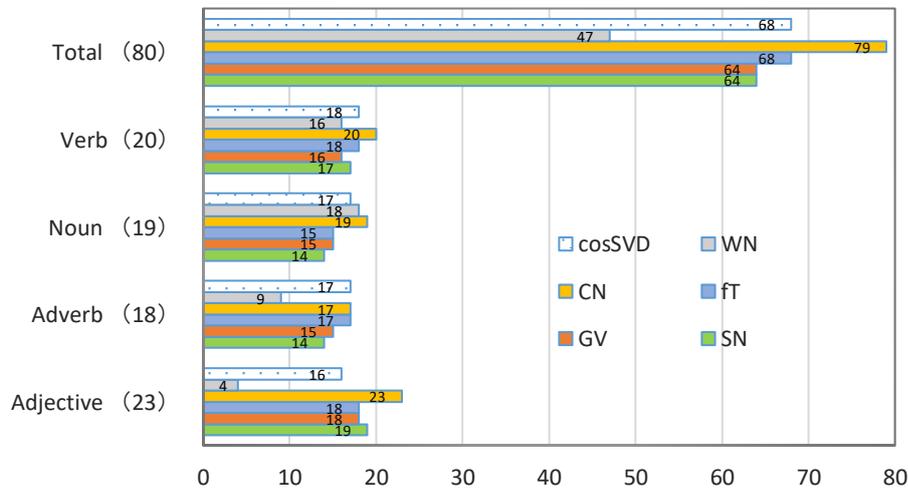

Figure 4. Comparison of different neural embeddings in detecting synonyms of TOEFL. The number of synonym questions under each subgroup is enclosed in the bracket beside each PoS tag. $cos_{SVD}$ denotes our dependency-constrained VSM. WN stands for the baseline for WordNet-based synonym detection

Both SN and GV attained an identical correctness rate of 80% in total, and fT arrived at a slightly better rate of 85%, equal to $cos_{SVD}$, our dependency-constrained VSM. However, BNC for extracting syntactic dependencies only consists of about 100 million tokens, much less than the English Wikipedia dump (about 2 billion tokens) for training the NNEs. SN, GV, and fT employ the same parameters and corpora in pre-training. Their comparable results on different subgroups of the synonym test indicated that although neural language models use different loss functions for semi-supervised learning and define different context contents in an n-gram of words (SN and GV) or characters (fT), they may function closely in calculating distributional semantics. Moreover, Levy et al. (2015) suggested that fine-tuning hyperparameters to maximise prediction probabilities in a neural language model may be a primary contributing factor for NNEs' superiority over the context-counting VSMs.

## 5. CONCLUSION

Recent NLP breakthroughs can be primarily contributed to neural language models, along with their byproduct of neural data representations such as SGNS and GloVe on words and BERT and GPT on sentences or documents. We systematically investigated neural word embeddings and dependency-constrained VSM in deducing distributional semantics. In the task of detecting synonyms in TOEFL, we achieved encouraging results compared to other syntactically conditioned and unconditioned context-counting VSMs, along with the context-predicting NNEs, which gained no particular advantage in computing distributional similarity notwithstanding their enormous size of training corpora. We can tentatively conclude that head-modifier and verb-object may bear semantic restrictions on verbs, and head-modifier dependencies may regulate the meaning of nouns. Although distributional semantics can mine the patterns of word usage in context, how to effectively inject human-compiled knowledge into VSMs may be a vital issue for future research on lexical semantics.

## ACKNOWLEDGEMENT

This research was supported by the Humanity and Social Science Foundation of China Ministry of Education (Grant No. 15YJA740054).